\def\year{2022}\relax
\documentclass[letterpaper]{article} 
\usepackage{aaai22}  
\usepackage{times}  
\usepackage{helvet}  
\usepackage{courier}  
\usepackage[hyphens]{url}  
\usepackage{graphicx} 
\usepackage{natbib}  
\usepackage{caption} 
\DeclareCaptionStyle{ruled}{labelfont=normalfont,labelsep=colon,strut=off} 
\usepackage{algorithm}
\usepackage{algorithmic}
\usepackage{epsfig}
\usepackage{amsmath}
\usepackage{amssymb}
\usepackage{bm}
\usepackage{subfigure}
\usepackage{booktabs}
\usepackage{multirow}
\usepackage{booktabs}
\usepackage{caption}
\usepackage{float}
\usepackage{algorithm, algorithmic}
\usepackage{bbding}
\usepackage{array}
\usepackage{mwe}
\usepackage{times}
\usepackage{epsfig}
\usepackage{graphicx}
\usepackage{amsmath}
\usepackage{amssymb}
\usepackage{newfloat}
\usepackage{listings}
\floatstyle{ruled}
\newfloat{listing}{tb}{lst}{}
\floatname{listing}{Listing}
\title{SiamTrans: Zero-Shot Multi-Frame Image Restoration 
\\ with Pre-Trained Siamese Transformers}
\author{Lin Liu\textsuperscript{1} \hspace{1mm}
Shanxin Yuan\textsuperscript{2}\thanks{Corresponding author} \hspace{1mm}
Jianzhuang Liu\textsuperscript{2}\hspace{1mm}
Xin Guo\textsuperscript{1} \hspace{1mm}
Youliang Yan\textsuperscript{2} \hspace{1mm}
Qi Tian\textsuperscript{3} \\
 \footnotesize{$^1$EEIS Department, University of Science and Technology of China}\\ \footnotesize{$^2$Huawei Noah's Ark Lab \qquad $^3$Huawei Cloud BU} \\ 
 \scriptsize{
   \{ll0825,willing\}mail.ustc.edu.cn  \{shanxin.yuan, liu.jianzhuang, yanyouliang, tian.qi1\}@huawei.com}
}
\usepackage{bibentry}
\begin{document}
\maketitle
\begin{abstract}
We propose a novel zero-shot multi-frame image restoration method for removing unwanted obstruction elements (such as rains, snow, and moir\'{e} patterns) that vary in successive frames.
It has three stages: transformer pre-training, zero-shot restoration, and hard patch refinement.
Using the pre-trained transformers, our model is able to tell the motion difference between the true image information and the obstructing elements.
For zero-shot image restoration, we design a novel model, termed SiamTrans, which is constructed by Siamese transformers, encoders, and decoders. Each transformer has a temporal attention layer and several self-attention layers, to capture both temporal and spatial information of multiple frames.
Only pre-trained (self-supervised) on the denoising task, SiamTrans is tested on three different low-level vision tasks (deraining, demoir\'{e}ing, and desnowing).
Compared with related methods, ours achieves the best performances, even outperforming those with supervised learning.
\end{abstract}

\section{Introduction}


Taking clean photographs under bad weather (\textit{e.g.}, snow and rain) or recovering clean images from occluding elements (\textit{e.g.}, moir\'{e} patterns), is challenging as the scene information is corrupted by these occluding elements in the captured images. These occluding elements can change quickly in a very short time (\textit{e.g.}, snow and rain) or due to the small movement of the camera (\textit{e.g.}, moir\'{e} patterns), making it difficult to do multi-frame image restoration.
%

Recovering the underlying clean image from a single degraded image is an ill-posed problem due to occlusions. Most of existing single image restoration methods that focus on dealing with these types of problem often mine high-level semantic information of the scene or the properties of degrading elements (\textit{e.g.}, noise and rain streak). But these single image restoration methods have difficulty in handling challenging and complex cases. 
To tackle these problems, multi-frame-based approaches, which use several images as input, are proposed to exploit additional information from supporting frames to the reference frame. 
Task-specific multi-frame-based methods include denoising~\cite{liang2020decoupled,mildenhall2018burst,godard2018deep}, demosaicing~\cite{ehret2019joint,kokkinos2019iterative}, super-resolution~\cite{farsiu2004fast,el2017new,wronski2019handheld, isobe2020video}, reflection removal~\cite{li2013exploiting,guo2014robust}, HDR imaging~\cite{dai2021wavelet, yan2019attention}, \textit{etc}. 
In addition, a few general frameworks have been proposed for multiple low-level vision tasks~\cite{xue2015computational,alayrac2019visual,liu2020learning,fan2020ScalewiseCF}.
The work in~\cite{alayrac2019visual} proposes a generic 3D CNN to estimate the foreground layer and \cite{liu2020learning} estimates the optical flows of obstruction elements. Both methods cannot obtain satisfactory results because it is hard to estimate either obstruction elements or their optical flows that vary dramatically in successive frames.
These supervised methods require a large amount of annotated training data and often have problems when applied to new tasks.

\begin{figure}[t!]
		\centering 
\includegraphics[width=0.43\textwidth]{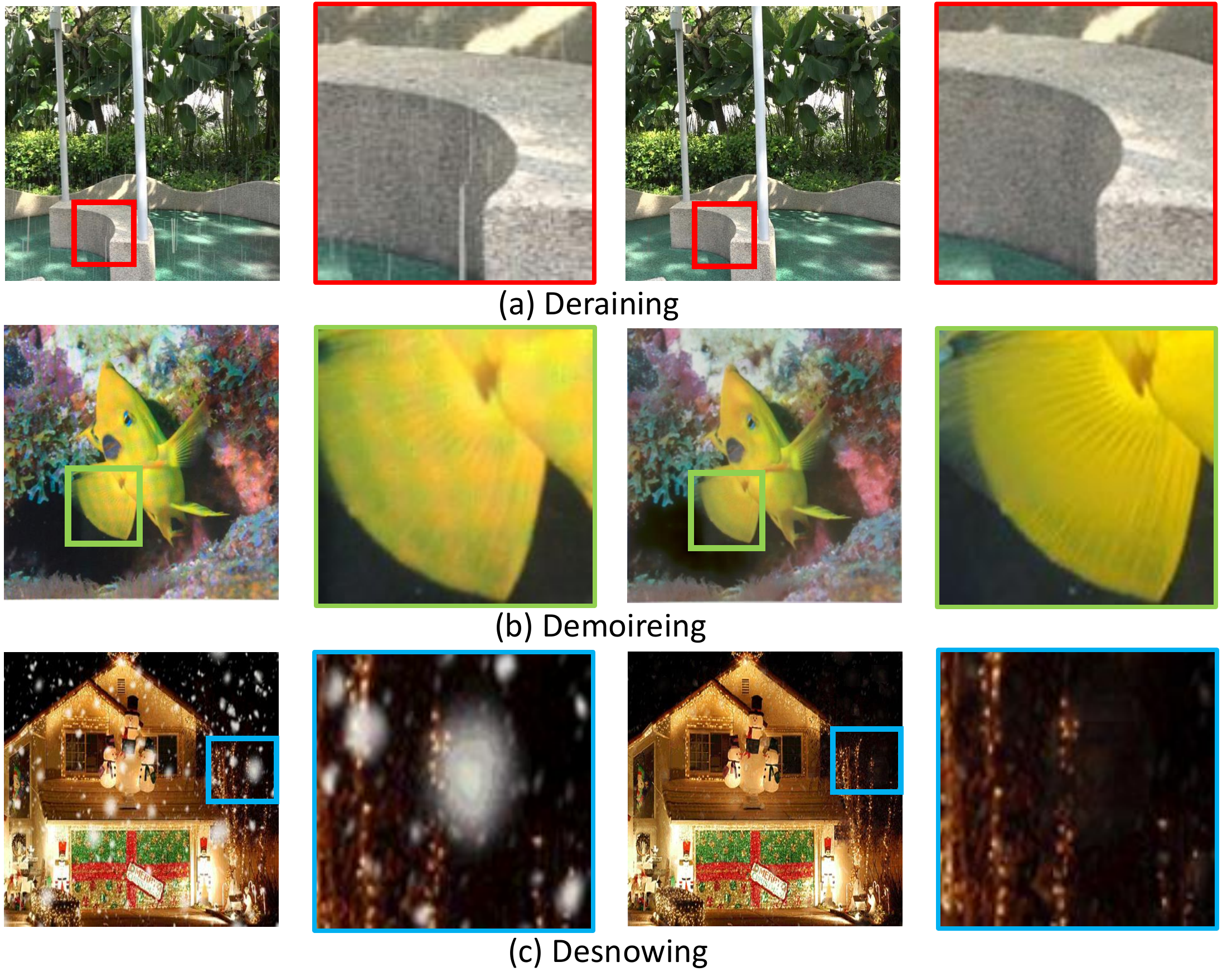}
     \vspace{-1mm}
    \caption{Example results of our zero-shot method for multi-frame deraining (top), demoir\'{e}ing (middle), and desnowing (bottom). Pre-trained on denoising only, our SiamTrans model removes unwanted elements while retaining the image details on multiple restoration tasks.}
    \label{fig:index}
\end{figure}

The pre-training and fine-tuning strategy is effective to obtain the natural image prior and adapt to new tasks. This strategy is often used on high-level vision tasks \cite{chen2020simple,grill2020bootstrap,he2020momentum} showing great performances, but it has not been widely used in low-level vision tasks yet.
Recently, some studies~\cite{gu2020image,pan2020exploiting,chan2020glean,bau2020semantic} use pre-trained GANs to do image restoration, where the GAN models are trained on large-scale natural images and can capture rich texture and shape priors. But the spatial information may not be faithfully kept due to low dimensionality of the latent code.
Transformer~\cite{vaswani2017attention} has been used in some low-level vision tasks very recently, such as image super-resolution~\cite{yang2020learning}, video synthesis~\cite{liu2020convtransformer} and video inpainting~\cite{zeng2020learning}. Image processing transformer (IPT)~\cite{chen2020pre} is pre-trained on three low-level vision tasks and outperforms state-of-the-art methods. It shows that the transformer is more advantageous than convolutional neural networks (CNNs) in large-scale data pre-training of low-level vision. However, IPT requires that the pre-training and fine-tuning are conducted on the same tasks, having difficulty in generalizing to completely unseen tasks, \textit{e.g.}, desnowing. Our model is a zero-shot learning setting, where the pre-training is only conducted on image denoising and the testing (without fine-tuning) is conducted on new tasks.

In this paper, inspired by the advantages of the multi-frame methods and pre-trained transformers, we propose a three-stage pipeline for multi-frame image restoration. This pipeline contains transformer pre-training, zero-shot restoration, and hard patch refinement.
In the first stage, the transformer is self-supervisedly pre-trained on the denoising task on a large scale dataset. The pre-training enables the transformer to learn natural image prior information between different frames and to converge fast in downstream iterations.
In the second stage, we design a model with Siamese Transformers (SiamTrans) for multiple downstream low-level tasks through zero-shot restoration. Note that the downstream tasks are unknown to the pre-training. SiamTrans consists of encoders, decoders, temporal attention modules, and self-attention modules. 
The aim of the third stage is to locate and refine hard-case patches.

In summary, we make the following contributions:
\begin{itemize}
  \item A three-stage pipeline for multi-frame image restoration is proposed. It consists of transformer pre-training, zero-shot restoration, and hard patch refinement. 
 
  \item We design a novel model with Siamese Transformers (SiamTrans) for zero-shot image restoration. Using pre-trained transformers with temporal and spatial attentions, our model is able to tell the motion difference between the nature image information and the obstructing elements. 
 
  \item When tested on three different low-level vision tasks (deraining, demoir\'{e}ing, and desnowing; see Fig.~\ref{fig:index}), our model achieves the best performances, even outperforming supervised learning methods.
\end{itemize}

\section{Related Work}

In this section, we present the most related works, including multi-frame image restoration, transformers and pre-training for low-level vision, and related tasks.

\textbf{Multi-frame image restoration.} Image restoration is an ill-posed problem and most of single image restoration methods~\cite{wang2018recovering,liu2019coherent,pan2020exploiting,zheng2020image,chen2020learning} often resort to high-level semantic information of the scenes or the degrading properties (\textit{e.g.}, noise level and rain streak). But these single image restoration methods have difficulty in handling challenging and complex cases.
To deal with these issues, multi-frame-based methods~\cite{godard2018deep,ehret2019joint,farsiu2004fast} have been proposed for low-level vision tasks, such as denoising~\cite{liang2020decoupled,mildenhall2018burst,godard2018deep}, demosaicing~\cite{ehret2019joint,kokkinos2019iterative}, super-resolution~\cite{farsiu2004fast,el2017new,wronski2019handheld} and reflection removal~\cite{li2013exploiting,guo2014robust}. 
In addition to these task-specific multi-frame methods, a few general frameworks have been proposed for multiple tasks~\cite{xue2015computational,alayrac2019visual,liu2020learning}.
Xue~\textit{et al.}~\cite{xue2015computational} present a computational approach for obstruction removal, which is applicable to multiple tasks, \textit{e.g.}, reflection removal and fence removal.
Liu~\textit{et al.}~\cite{liu2020learning} estimate dense optical flow fields of the background and degrading element layers and then reconstruct them. Our work is an unsupervised zero-shot multi-frame image restoration method that can be applied to multiple low-level vision tasks.

\textbf{Transformers for low-level vision tasks.} Transformers~\cite{vaswani2017attention} are a neural network framework using the self-attention mechanism. They are originally used in natural language processing, and then used in computer vision tasks including low-level vision very recently~\cite{yang2020learning,chen2020pre,zeng2020learning}. 
Yang \textit{et al.}~\cite{yang2020learning} propose a texture transformer network for image super-resolution. It transfers relevant textures from reference images to low-resolution images.
Chen \textit{et al.}~\cite{chen2020pre} develop a pre-trained transformer called IPT for three low-level vision tasks, which outperforms state-of-the-art methods.
In low-level video processing, Liu \textit{et al.}~\cite{liu2020convtransformer} propose ConvTransformer to synthesize video frames. 
Zeng \textit{et al.}~\cite{zeng2020learning} propose a spatial-temporal transformer network for video inpainting, where frames with holes are taken as input and the holes are filled simultaneously.

\textbf{Pre-training for low-level vision tasks.}
The pre-training and fine-tuning strategy is often used on high-level vision tasks, showing good performances~\cite{chen2020simple,grill2020bootstrap,he2020momentum}. For low-level vision tasks, the random initialization and end-to-end training strategy is usually adopted.
More recently, some studies~\cite{gu2020image,pan2020exploiting,chan2020glean,bau2020semantic} use pre-trained GANs to do image restoration. 
A GAN model trained on a large-scale set of natural images can capture rich texture and shape priors.
The problem of using pre-trained GANs for image restoration is that some spatial details may not be faithfully recovered due to the low dimensionality of the latent code, resulting in artifacts compared with ground-truth.
Recently, IPT~\cite{chen2020pre} shows that transformers are more advantageous than CNNs in large-scale data pre-training for low-level vision. %
IPT uses task-specific embeddings as an additional input for the decoder, where the task-specific embeddings are learned to decode features for different tasks. 
Different from IPT, our SiamTrans is pre-trained for zero-shot image restoration. 
We only need to pre-train it on the denoising task and then apply it to multiple downstream low-level tasks which are unknown to the pre-training.

\textbf{Deraining, desnowing, and demoir\'{e}ing.}
Deraining methods can be grouped into single image derainging and video deraining.
The former group focuses on mining the intrinsic prior of the rain signal~\cite{fu2017removing,yang2017RainRemoval,deng2018directional,li2018recurrent,guo2020EfficientDeRainLP}. 
Compared with single-image rain removal, video deraining utilizes temporal information to detect and remove rains~\cite{jiang2019fastderain,li2019video,yang_2020_CVPR,li2021OnlineRR}.
We take advantage of transformer's ability to acquire natural image prior and temporal information after training on a large scale dataset.

Compared with rain, snow is more complicated due to its large variations of size and shape, and its transparency property. Snow removal methods also include single image desnowing~\cite{chen2020jstasr,jaw2020desnowgan,liu2018desnownet} and video desnowing~\cite{ren2017video,li2019video}. 
For single image desnowing,  Liu~\textit{et al.}~\cite{liu2018desnownet} propose a learning based model. Chen~\textit{et al.}~\cite{chen2020jstasr} design a desnowing network which contains three parts: snow removal, veiling effect removal, and clean image discriminator.
For video desnowing, Ren~\textit{et al.}~\cite{ren2017video} use matrix decomposition to desnow.

Moiré artifacts are not unusual in digital photography, especially when photos are taken of digital screens. Moiré patterns are mainly caused by the interference between the screen’s subpixel layout and the camera’s color filter array. 
Recently, some deep learning models~\cite{hefhde2net,zheng2020image,yang2020high,he2019mop,liu2020wavelet,liu2020self,zheng2021learning,yuan2019aim,yuan2019aims,yuan2020ntire} are proposed for single image demoiréing. For multi-frame demoiréing, Liu~\textit{et al.}~\cite{liu2020mmdm} use multiple images as inputs and design multi-scale feature encoding modules to enhance low-frequency information. 
Unlike their approach, our method is unsupervised and does not need to train with a large number of moiré and moiré-free image pairs. We only need pre-training on denoising and then do zero-shot restoration with multiple moiré frames.

\begin{figure*}[t]
		\centering 
		\includegraphics[width=0.92\textwidth]{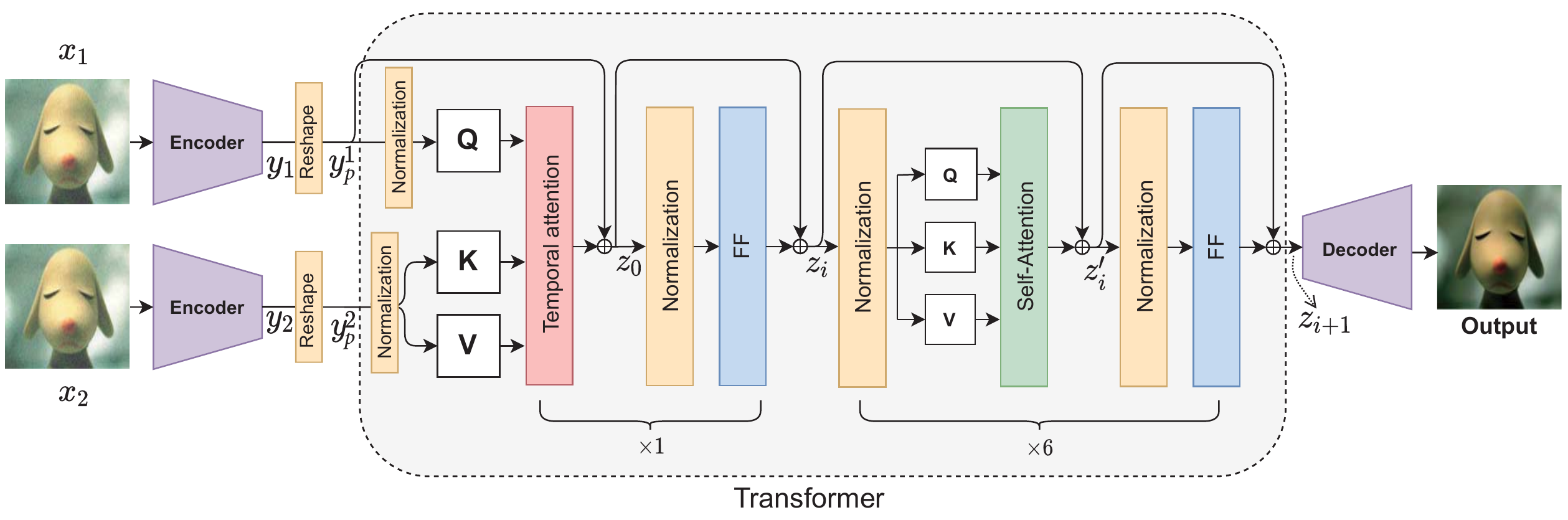}
		\caption{The architecture of our basic network, which consists of three parts: two CNN encoders, a transformer with both temporal and spatial attention modules, and a CNN decoder. Note that we also add learnable position embeddings~\cite{dosovitskiy2021image} to the input sequence of the transformer, which are not displayed here.}
		\label{network_train}
\end{figure*}	

\section{Proposed Method}
In this section, we first introduce our basic network architecture and then present the three stages of our method, including transformer pre-training, zero-shot restoration with SiamTrans, and hard patch refinement.
This basic network is the building block of our SiamTrans.

\subsection{Basic Network Architecture}
\label{sec:basicnetwork}
As shown in Fig. \ref{network_train}, our basic network includes two weight-sharing CNN encoders each corresponding to an input, and a CNN decoder to generate the final output. Between the encoders and the decoder, we construct a transformer that has a temporal attention module and six spatial self-attention modules. 

\textbf{Temporal attention module.}
In multi-frame image restoration, information from supporting frames can help to restore the corrupted reference frame. Given the input images $x_{1}$ and $x_{k}$ $(k\in\{2, \ldots, N\})$, we present the process of the encoding as:
\begin{equation}
    y_{1} = E(x_{1}),~~y_{2} = E(x_{k}), 
    \label{eq1}
\end{equation}
where $y_{1},\ y_{2} \in \mathbb{R}^{C \times H \times W}$ denote the output feature maps of the encoders, $C$ is the number of feature channels, $H$ and $W$ are the height and width of the feature map, respectively.

After feature extraction, the obtained feature maps, $y_{j},$ $j\in \{1, 2\}$ is reshaped into a sequence of flattened patches (vectors), $y^{j}_{p}=\{ y^{j}_{p_{1}}, y^{j}_{p_{2}}, \ldots, y^{j}_{p_{m}} \} ,$ where $y^{j}_{p_{i}} \in \mathbb{R}^{CP^{2} }, i\in\{1, \ldots, m\}$;
$m=\frac{H W}{P^{2}}$ is the total number of patches and $P \times P$ is the patch size. 
The process of the temporal attention module is formulated as:
\begin{equation}
z_{0}=(\mathrm{MHA}\left(\mathrm{NL}(y^{1}_{p}), \mathrm{NL}(y^{2}_{p}), \mathrm{NL}(y^{2}_{p})\right) + y^{1}_{p},\\
\end{equation}
\begin{equation}
z_{1}=\mathrm{FF}\left(\mathrm{NL}(z_{0})\right)+z_{0},
\end{equation}
where $\mathrm{MHA}(Q,K,V)$ denotes the  multi-head attention module with $Q=\mathrm{NL}(y^{1}_{p})$, $K=\mathrm{NL}(y^{2}_{p})$, and $V=\mathrm{NL}(y^{2}_{p})$ corresponding to the three basic transformer elements Query, Key, and Value, respectively, NL denotes the operation of the normalization layer, and FF is a feed forward network~\cite{dosovitskiy2021image,vaswani2017attention}.

\textbf{Self-attention module \& decoder.}
After the fusion of two corresponding frames by the temporal attention module, the self-attention module uses the self-attention mechanism to extract useful spatial information from $z_{i}$. %
In our work, six self-attention modules are employed, each with a multi-head self-attention layer and a feed forward network. The process is represented as:
\begin{equation}
z_{i}^{\prime}=(\mathrm{MHA}\left(\mathrm{NL}(z_{i}), \mathrm{NL}(z_{i}), \mathrm{NL}(z_{i})\right) + z_{i}, i=1,2,...,6,\\
\end{equation}
\begin{equation}
z_{i+1}=\mathrm{FF}\left(\mathrm{NL}(z_{i}^{\prime})\right)+z_{i}, i=1,2,...,6.
\end{equation}
Finally the output $z_{7}$ is reshaped to $g \in \mathbb{R}^{C \times H \times W}$.
The output of the decoder is $o=D(g) \in \mathbb{R}^{3 \times H \times W}$.

\subsection{Transformer Pre-Training}

We pre-train the basic network in Fig. \ref{network_train} such that it can capture the intrinsic properties and transformations of various images, \textit{i.e.}, image prior. The pre-training task is denoising with the Place365 dataset~\cite{zhou2017places}. 
We choose 328,000 images for the pre-training. In each iteration, we randomly choose an image $I$ and a noise level $ \sigma \in [ 1,50 ] $ and synthesize two degraded images by:
\begin{equation}
    x_{1} = I +\mathcal{N}_{1},~~x_{2} = I +\mathcal{N}_{2}, 
    \label{eq5}
\end{equation}
where $\mathcal{N}_{1}$ and $\mathcal{N}_{2}$ are two different samples from the Gaussian noise distribution $\mathcal{N}(0,\sigma)$. The loss function in the pre-training stage is,
\begin{equation}
L_{\text{pre-train}}=\left\|M(x_{1},x_{2})- I\right\|_{1},
\end{equation}
where $M$ denotes the basic network. After the pre-training, the network learns the feature correlation of different frames and the natural image prior. The ablation study in Sec. \ref{sec:exp} shows that the pre-training cannot only make SiamTrans converge faster but also improve its performance. 

\subsection{SiamTrans for Zero-Shot Restoration}
\label{sec:siamtrans}

\begin{figure}[!t]
		\centering
		\includegraphics[ width=0.48\textwidth]{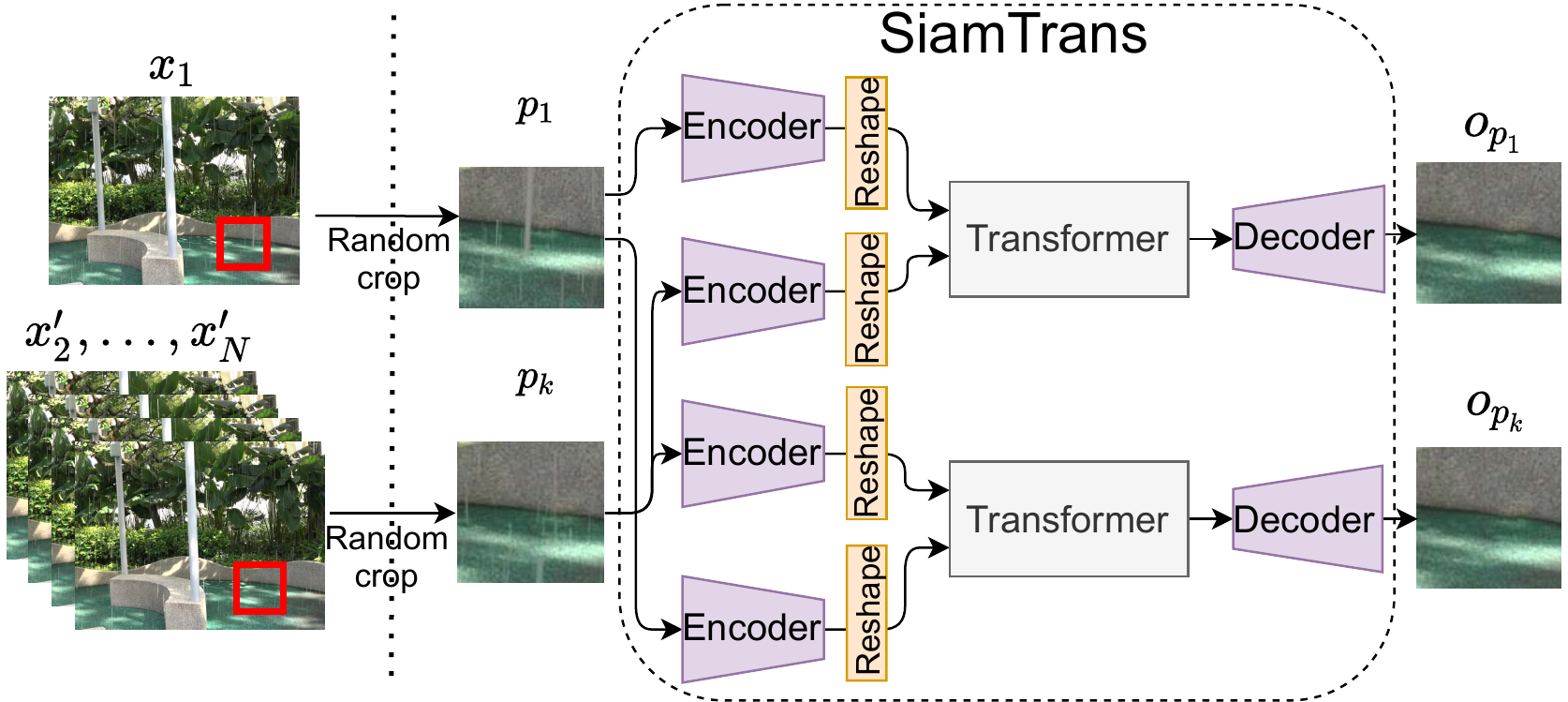}
		\caption{The architecture of our SiamTrans for zero-shot restoration. It includes Siamese transformers, four encoders and two decoders. $x_{2}^{'},...,x_{N}^{'}$ are obtained by warping $x_{2},...,x_{N}$ towards $x_{1}$. Two same-location patches $p_1$ and $p_k, \ k \in \{2,3,...,N\}$, are randomly cropped and served as the inputs to SiamTrans.}  %
		\label{network_finetue}
\end{figure}

The SiamTrans model is shown in Fig. \ref{network_finetue}, which is formed by two basic networks, with four weight-sharing CNN encoders, two weight-sharing transformers and two weight-sharing CNN decoders.
Suppose we have a short sequence of images $\{x_{1},x_{2},...,x_{N}\}$, where $x_{1}$ is the reference frame. The task of multi-frame image restoration is to recover a clean image $o_{1}$ corresponding to $x_{1}$. 
The images $\{x_{2},...,x_{N}\}$ are first warped to $x_{1}$ by FlowNet~\cite{ilg2017flownet}, resulting in the aligned images $\{x'_{2},...,x'_{N}\}$. 
In each iteration, we randomly crop the patches $p_{1}$ and $p_{k}$ ($\ k \in \{2,3,...,N\}$) of the same location from $x_{1}$ and $x'_{k}$ respectively.
$p_{1}$ and $p_{k}$ are then sent to SiamTrans.
We define the loss function for the restoration as,
\begin{eqnarray}
\begin{split}
L_{\text {zero-shot}}=&\left\|M_{1}\!\left(p_{1},p_{k}\right)\!-\!M_{2}\!\left(p_{k}, p_{1}\right)\right\|_{1} \\&
+\lambda \left(\left\|M_{1}\!\left(p_{1}, p_{k}\!\right)\!-\!p_{1}\right\|_{1} 
\!+\! \left\|M_{2}\!\left(p_{k}, p_{1}\!\right)\!-\!p_{k}\right\|_{1}\right),\\
\end{split}
\label{eq10}
\end{eqnarray}
where the first term is the consistency loss, and the second term is the fidelity loss, and $o_{p_{1}}=M_{1}(p_{1},p_{k})$ and $o_{p_{k}}=M_{2}(p_{k},p_{1})$ are the outputs of the two basic networks $M_{1}$ and $M_{2}$, respectively.

After a number of iterations with different random patch pairs, SiamTrans has learnt how to restore the clean images from $x_{1},x^{\prime}_{2},...,x^{\prime}_{N}$. Then, we use $M_{1}$ or $M_{2}$ to obtain the restored images $o_{1}$, $o_{2},...,$ and $o_{N}$ from $x_{1}$, $x^{\prime}_{2},...,$ and $x^{\prime}_{N}$, respectively.

\subsection{Hard Patch Refinement}
\label{sec:hpr}
\begin{figure}[!t]
		\centering
		\includegraphics[width=0.48\textwidth]{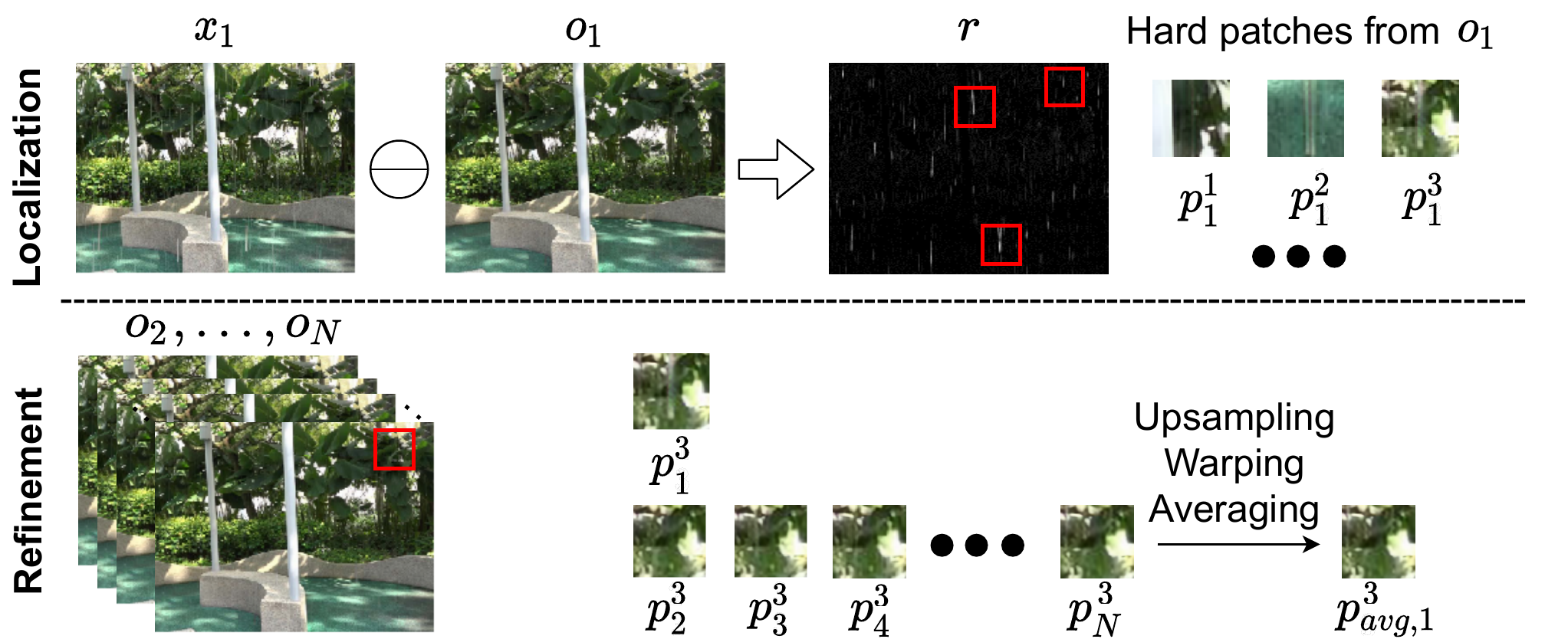}
		\caption{Procedure of our hard patch refinement, where one example with patch $p^{3}_{1}$ is given.}
		\label{fig:hpr}
\end{figure}
After the initial restoration described in Sec. \ref{sec:siamtrans}, SiamTrans with learned image prior can recover a good result for a specific scene.
However, due to the variety of degradation, it is difficult to get a completely clean image in the zero-shot setting.
So we design a hard patch refinement to locate and recover the patches where the degradation has not been well tackled. 

\textbf{Localization.} 
As shown in Fig. \ref{fig:hpr}, to localize the hard patches in the frame $x_{1}$, we generate a residual map $r = \left\|o_{1}-x_{1}\right\|_{1}$.
On the residual map $r$, we select $n$ points with the highest values, where the patches of size $s \times s$ centered at these points are non-overlapping.

\textbf{Refinement.}
After locating the hard patches that are not well restored, we extract $n$ $s \times s$ patches ${p^{1}_{j},p^{2}_{j},...,p^{n}_{j}}$ at the $n$ centers of $o_{j},\  j=1,2,..., N$.
We update each patch in each frame iteratively.
For every patch $p^{k}_{1}$ in $o_{1}$, we update $p^{k}_{1}$ as follows:
\begin{equation}
\begin{split}
    p^{k}_{1} \leftarrow \  &\alpha \times p^{k}_{1} + \frac{ 1 - \alpha}{N-1} \!\left(\sum_{m=2}^{N} \!W_{1}(p^{k}_{m}\!\uparrow\!)\right)\!\downarrow,
\end{split}
\label{eq11}
\end{equation}
where $\alpha$, $W_{1}$, $\uparrow$, and $\downarrow$ denote the balancing parameter, warping towards $o_{1}$, upsampling, and downsampling, respectively. 
We use the pre-trained LIIF model~\cite{chen2020learning} to perform upsampling and downsampling. The upsampling is for better alignment of the frames.
Then, for every patch $p^{k}_{j}$ in $o_{j},\  j=2,3,..., N$, we update $p^{k}_{j}$ as follows:

\begin{equation}
\begin{split}
    p^{k}_{j} \leftarrow \ &\alpha \times p^{k}_{j} + \frac{ 1 - \alpha}{N-2} \!\left(\sum_{m=2, m\neq j}^{N} \!W_{j}(p^{k}_{m}\!\uparrow\!)\right)\!\downarrow,
\end{split}
\label{eq12}
\end{equation}
where $W_{j}$ denotes warping towards $o_{j}$.
\section{Experiments and Analysis}
\label{sec:exp}
In this section, we show ablation study and comparison with state-of-the-art methods.
Our algorithm is implemented on a NVIDIA Tesla V100 GPU in PyTorch. The network is optimized with the Adam~\cite{kingma2014adam} optimizer.
In both the pre-training and zero-shot restoration, the batch size is set to 1 and the initial learning rate is $1 \times 10^{-5}$. 
The algorithm runs for 20 epochs and 20 iterations for the pre-training and the hard patch refinement, respectively.
For zero-shot restoration, it takes 200, 500, and 1000 iterations for demoir\'{e}ing, desnowing, and deraining, respectively. 
The $\lambda$ and $\alpha$ in Eqn. \ref{eq10} and Eqn. \ref{eq11}/Eqn. \ref{eq12} are empirically set to 5 and 0.9 respectively.
Besides, the feature map size $H \times W$ in Sec. ~\ref{sec:basicnetwork} is $128 \times 128$, the patch size  $p \times p$ in Sec. ~\ref{sec:basicnetwork} is $32 \times 32$, the patch size $s \times s$ in Sec. ~\ref{sec:hpr} is also $32 \times 32$, and the patch number $n$ in Sec. ~\ref{sec:hpr} is 50.
The structures of the CNN encoders and decoders can be found from the supplementary materials.

 \begin{figure*}[!t]
 		\centering
 		\includegraphics[width=0.90\textwidth]{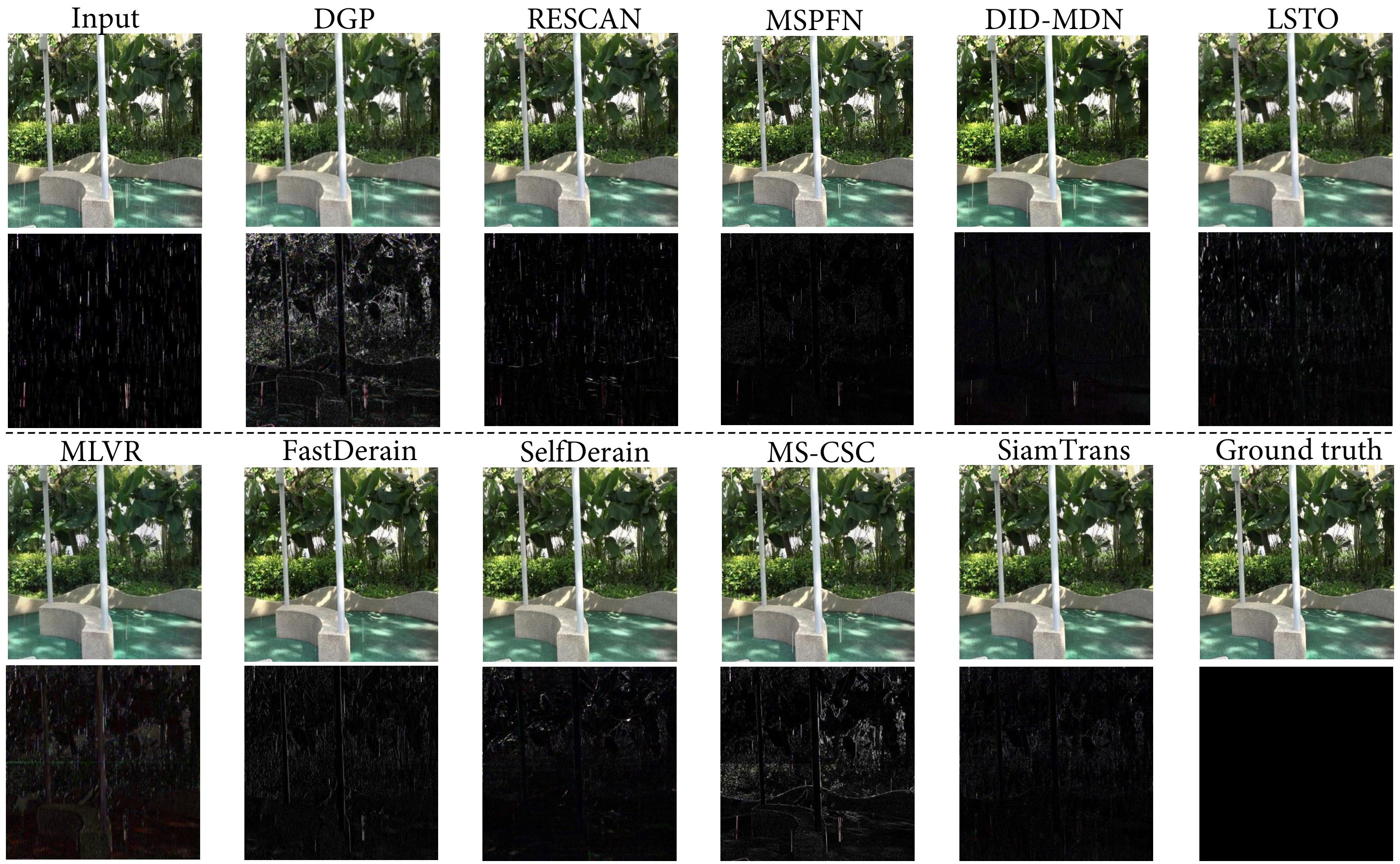}
 		\caption{Visual deraining comparison among our method and other methods including deraining-specific methods (RESCAN, MSPFN, DID-MDN, FastDerain, SelfDerain, and MS-CSC) and general restoration methods (DGP, LSTO, and MLVR),  evaluated on \textit{NTURainSyn}. The second and the last rows are the differences between the predicted images and the ground truth.}
 		\label{fig:rain}
 \end{figure*}
 
\subsection{Datasets and State-of-the-Arts}
\label{sec:dataset}
\textbf{Datasets.} 1) Deraining. Since there is no existing short-sequence deraining dataset, we build our multi-frame deraining test set through extracting adjacent frames from the NTURain dataset~\cite{chen2018robust}, where the images are taken from an unstable panning camera with slow movements.
In total, we extract 40 synthetic rain sequences (denoted as \textit{NTURainSyn}) and 12 real rain sequences (denoted as \textit{NTURainReal}), where each sequence contains 8 consecutive frames.
The training set for the compared supervised methods is Rain100L~\cite{yang2017RainRemoval}, which contains 1800 scenes and rain maps\footnote{The detailed information about how to synthesize rain sequences is described in the supplementary materials.}.

2) Demoir\'{e}ing. We create a multi-frame demoir\'{e}ing dataset (\textit{MFMoir\'{e}}) to evaluate our method quantitatively and qualitatively. 
We collect 146 high quality images from the Internet as ground truth. To get pre-aligned moir\'{e} sequences, we adopt the method in~\cite{sun2018moire} to align the ground truth and corresponding moir\'{e} images.
Each moir\'{e} sequence contains 10 moir\'{e} images. Note that the moir\'{e} patterns vary a lot within a sequence and across different sequences.
We split the 146 sequences into the training set with 116 sequences (for compared supervised methods) and the testing set with 30 sequences.

3) Desnowing. We create a multi-frame desnowing dataset (\textit{MFSnow}) to evaluate our method.
We use the ground-truth images in \textit{NTURainSyn} as snow-free images and synthesize corresponding snow frames using the method in \cite{liu2018desnownet}. Finally, 3000 training snow sequences (for compared supervised methods) and 30 testing snow sequences are collected.

 \begin{figure*}[t]
 		\centering
 		\includegraphics[width=0.82\textwidth]{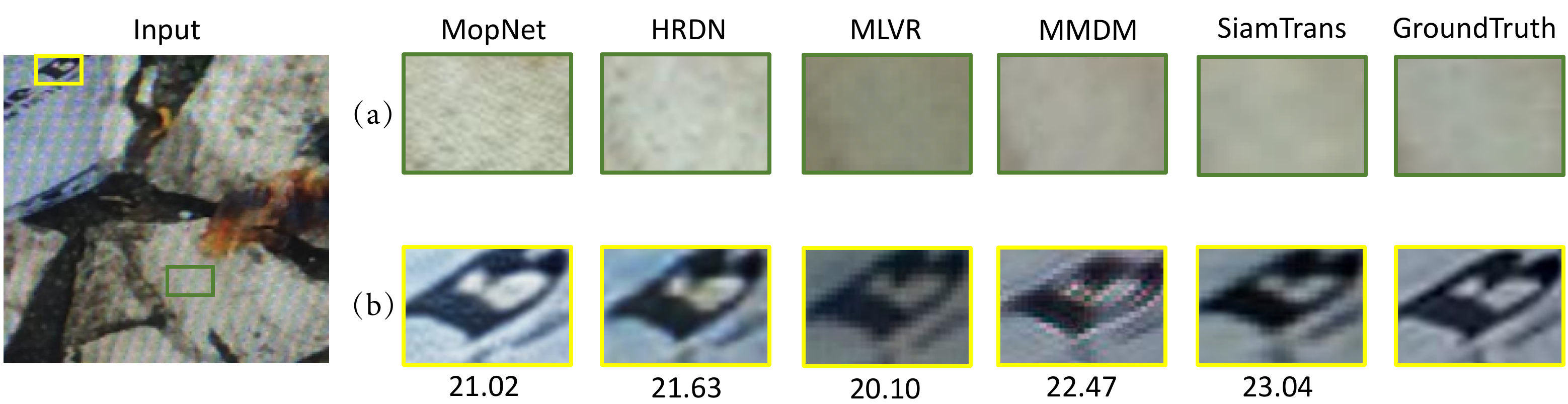}
 		\caption{Visual demoir\'{e}ing comparison among our method and other algorithms including three demoir\'{e}ing-specific methods (MopNet, HRDN and MMDM) and a general image restoration method (MLVR),  evaluated on \textit{MFMoir\'{e}}. The numbers at the bottom of the patches are the PSNRs of the corresponding methods for the whole restored images.}
 		\label{fig:moire}
 \end{figure*}

 \begin{table}[t]
  \begin{center}
  \caption{Quantitative deraining comparison. The best results are in \textbf{bold}. (a): \textit{NTURainSyn}; (b): \textit{NTURainReal}. }
  \label{tab:nturain}
  \resizebox{8.5cm}{!}{
  \begin{tabular}{cccccc}
    \toprule
       &Method &DGP~\cite{pan2020exploiting}  &MSPFN~\cite{Kui_2020_CVPR}&MS-CSC~\cite{li2019video}& FastDeRain~\cite{jiang2019fastderain}\\
    \midrule 
    \multirow{2}*{(a)}&PSNR$\uparrow$&20.67 &25.16&24.78&25.75\\
    &SSIM$\uparrow$& 0.5291&0.8497&0.7344&0.8991\\
    \midrule 
    (b)&NIQE$\downarrow$&4.051 &3.462&3.368&3.627\\
    \midrule 
    \midrule 
     &Method &SelfDeRain~\cite{yang_2020_CVPR}&MLVR~\cite{alayrac2019visual} &LSTO~\cite{liu2020learning} & SiamTrans (Ours)\\
    \midrule 
    \multirow{2}*{(a)}&PSNR$\uparrow$&26.81&26.78 & 24.72& \textbf{27.02}\\
    &SSIM$\uparrow$& 0.8935&0.8678& 0.8853&\textbf{0.9024}\\
    \midrule 
    (b)&NIQE$\downarrow$&3.322&3.316 & 3.354& \textbf{3.302} \\
    \bottomrule 
  \end{tabular}}
  \end{center}
\end{table}

\begin{table}[t]
  \begin{center}
  \caption{Quantitative demoir\'{e}ing and desnowing comparison. The best results are in \textbf{bold}. }
  \label{tab:mfmoire}
  \resizebox{7.0cm}{!}{
  \begin{tabular}{ccccc }
    \toprule
      &Method & PSNR$\uparrow$ & SSIM$\uparrow$ &LPIPS$\downarrow$\\
    \midrule 
    \multirow{4}*{MFMoir\'{e}}&LSTO~\cite{liu2020learning}&21.51&0.6504& 0.3573\\
    &MLVR~\cite{alayrac2019visual}&19.87&0.5904& 0.5022\\
    &MMDM~\cite{liu2020mmdm}&21.61&0.6476 &0.3710\\
    &SiamTrans (Ours)&$\mathbf{22.26}$&$\mathbf{0.6642}$ &$\mathbf{0.3197}$ \\
    \midrule
    \multirow{4}*{MFSnow}&LSTO~\cite{liu2020learning}&23.41&0.8228&0.2056 \\
    &MLVR~\cite{alayrac2019visual}&21.88&0.8064&0.2233 \\
    &MS-CSC~\cite{li2019video}&23.16&0.8201 &0.2137\\
    &OTMSCSC~\cite{li2021OnlineRR}&24.21&0.8332 &--\\
    &SiamTrans (Ours)&$\mathbf{26.05}$&$\mathbf{0.8605}$ & $\mathbf{0.1323}$\\
    \bottomrule
  \end{tabular}}
  \end{center}
\end{table}

\begin{table}[t]
  \centering\small
  \caption{Ablation study on \textit{MFMoir\'{e}}. }
  \vspace{-2mm}
  \resizebox{6.6cm}{!} {
  \begin{tabular}{c|ccc}
    \toprule
      Model&PSNR$\uparrow$&SSIM$\uparrow$&LPIPS$\downarrow$\\
      \midrule
      W/o pre-training&18.92 &0.535 &0.6248 \\
      W/o Hard Patch Refinement& 22.23&0.663 &0.3244 \\
      W/o TA Modules&21.68 &0.653 &0.3402 \\
      two SA Modules& 21.87&0.658 &0.3387 \\
      four SA Modules&22.01 &0.661 & 0.3270\\
      UNet& 21.56&0.649 &0.3570 \\
      SiamTrans& $\mathbf{22.26}$&$\mathbf{0.664}$ &$\mathbf{0.3197}$\\
    \bottomrule
    
  \end{tabular}}
  \vspace{-2mm}
  \label{tab:abstudy1}
\end{table}

\textbf{State-of-the-art methods.} 1) For multi-frame deraining, we compare with nine state-of-the-art methods, including three supervised single-image deraining methods (RESCAN~\cite{li2018recurrent}, MSPFN~\cite{Kui_2020_CVPR}, and DID-MDN~\cite{derain_zhang_2018}), one unsupervised image restoration method (DGP~\cite{pan2020exploiting}), three unsupervised video deraining methods (MS-CSC~\cite{li2019video}, FastDerain~\cite{jiang2019fastderain}, and SelfDerain~\cite{yang_2020_CVPR}), and two supervised multi-frame image restoration methods (MLVR~\cite{alayrac2019visual} and LSTO~\cite{liu2020learning}). 
2) For multi-frame demoir\'{e}ing, we compare with five state-of-the-art methods, including two supervised single-image demoir\'{e}ing methods (MopNet~\cite{he2019mop} and HRDN~\cite{yang2020high}) and three multi-frame demoir\'{e}ing methods (MMDM~\cite{liu2020mmdm}, MLVR, and LSTO).
3) For multi-frame desnowing, we compare with one single-frame desnowing method (JSTASR~\cite{chen2020jstasr}) and three video desnowing methods (MS-CSC, MLVR, and LSTO). Except MS-CSC, the other three desnowing methods are supervised.
For all the methods, we use their default parameters to generate the results.

\subsection{Comparison with State-of-the-Arts.}
\textbf{Quantitative results.} 
On the datasets with ground truth (\textit{e.g.}, \textit{NTURainSyn} and \textit{MFMoir\'{e}}), we use PSNR, SSIM~\cite{wang2004image}, and Learned Perceptual Image Patch Similar (LPIPS)~\cite{zhang2018perceptual} to compare the restored images. LPIPS measures perceptual image similarity using a pre-trained deep network.
On the dataset without ground truth (\textit{NTURainReal}), we evaluate all generated images using a no-reference quality metric, NIQE~\cite{mittal2012making}.
As shown in Table \ref{tab:nturain} and Table \ref{tab:mfmoire}, the proposed method obtains the best scores on all the evaluation metrics and on all the datasets.
More details about the comparison can be found from the supplementary materials.

\textbf{Qualitative results.}
As shown in Fig. \ref{fig:rain}, the first and third rows are the predicted results except the input and the ground truth. The second and the last rows are the differences between the predictions and the ground truth. Through a pre-trained model, the output of DGP losses many image details. The single-image deraining methods, RESCAN, MSPFN and DID-MDN, cannot remove the rain streaks thoroughly because of the limitation of their generalization ability. Our method removes the rain streaks and retains the image details at the same time.

For image demoir\'{e}ing, as shown in Fig. \ref{fig:moire}(a), the output results of the single-image demoir\'{e}ing methods (MopNet and HRDN) may keep some noise or moir\'{e} artifacts. Fig. \ref{fig:moire}(b) shows that the multi-frame methods, MLVR and MMDM, face the color-shift and color artifact problems, respectively. The visual results of desnowing are shown in the supplementary materials.

\subsection{Ablation Study}

To illustrate the contributions of the three stages in our pipeline, we conduct ablation study on \textit{MFMoir\'{e}}.
The quantitative comparisons are shown in Table \ref{tab:abstudy1}, where the first row are different variants of our model described below.

\subsubsection{Importance of the stages.}
Although the second stage (zero-shot restoration) is the core of our pipeline, the first and third stages are also important.

\begin{figure}[t]
		\centering
		\includegraphics[width=0.46\textwidth]{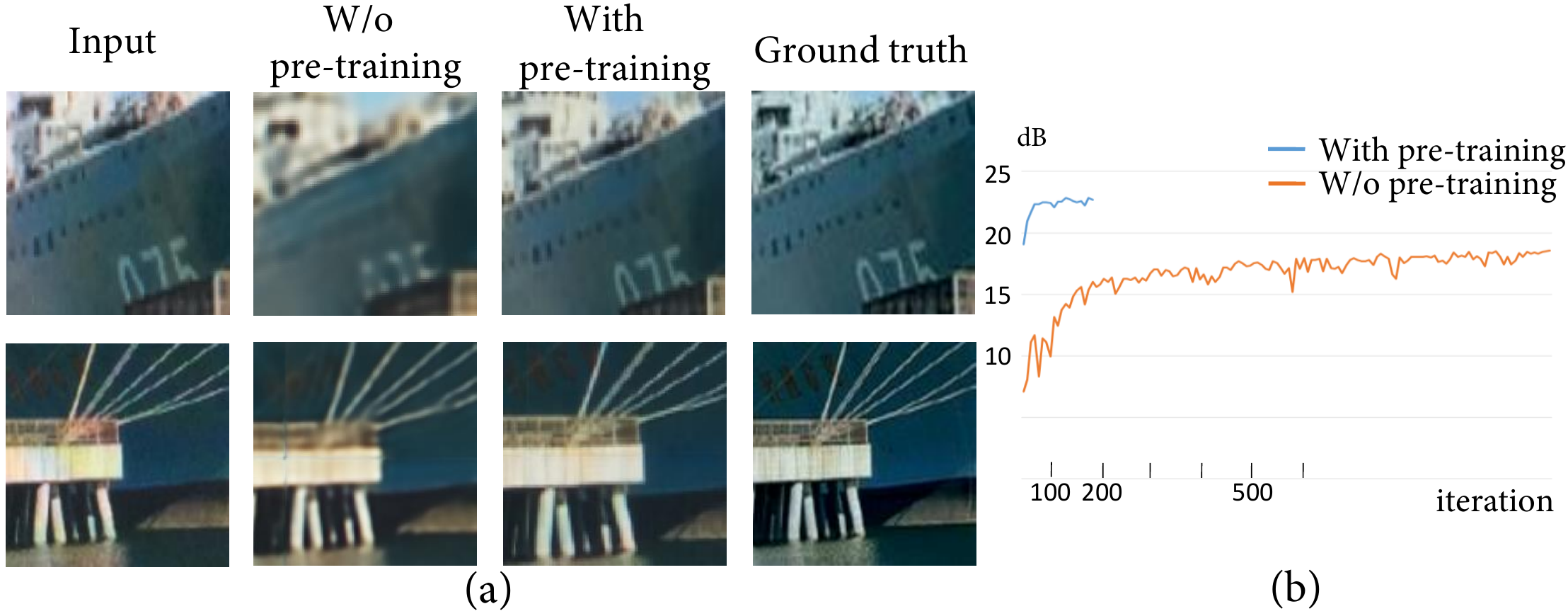}
		\vspace{-1mm}
		\caption{Demoir\'{e}ing comparison between our two models with and without the pre-training stage.}
		\label{fig:abstudy_pretrain}
\end{figure}

\textbf{Pre-training.}
The model of `W/o pre-training' means the model is randomly initialized without going through the pre-training stage. Compared with the final full model `SiamTrans', it shows that removing the pre-training stage leads to significant performance drop of 3.34dB on PNSR. 
As shown in Fig.~\ref{fig:abstudy_pretrain}(a), the results of `W/o pre-training' are blurry and lose some details.
These results indicate that the pre-training is important for the later stages. Without the pre-training, even if we train the later stages longer (\textit{e.g.}, increasing the number of iterations for zero-shot restoration ten times to 2000), the network still cannot capture the details.
Fig.~\ref{fig:abstudy_pretrain}(b) illustrates the numbers of iterations for both models (with and without the pre-training) needed to reach convergence for the zero-shot restoration stage. %
It clearly shows that the pre-training enables the network to converge much faster and perform much better in this stage.

\begin{figure}[t]
		\centering
		\includegraphics[width=0.45\textwidth]{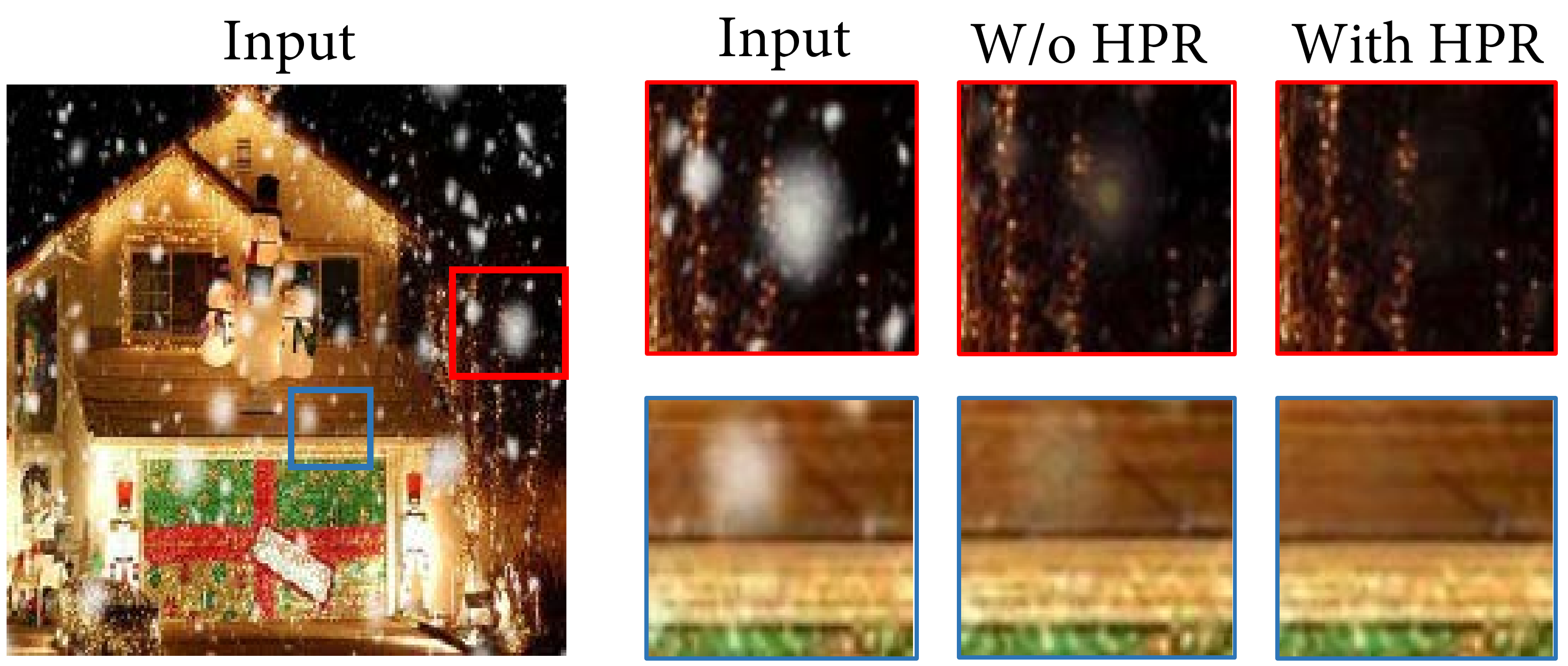}
		\vspace{-1mm}
		\caption{One example of the desnowing results of our method with HPR and without HPR.}
		\label{fig:hpr_abstudy}
\end{figure}

\textbf{Hard patch refinement (HPR).}
To verify the contribution of the HPR, we remove the 3rd stage from the whole pipeline (indicated as `W/o HPR' in Table ~\ref{tab:abstudy1}). 
Although its performance drop is not as serious as `W/o pre-training' in terms of the metrics in Table~\ref{tab:abstudy1}, its visual quality is degraded obviously in many cases.
As shown in Fig. \ref{fig:hpr_abstudy}, SiamTrans can better recover the image occluded by the snow. 

\vspace{-2mm}

\subsubsection{The Network Architecture.}

We also conduct a study to analyze the effectiveness of the network architecture.

\textbf{Temporal  attention  module \& self-attention module.}
In Table \ref{tab:abstudy1}, `W/o TA' means that in the basic network, the temporal-attention module and one encoder are removed and the input is a single frame. 
The performance drop in Table~\ref{tab:abstudy1} verifies the necessity of the temporal-attention module.
The self-attention module in the transformer can be stacked to enhance the learning ability. We verify the effectiveness of using multiple self-attention modules.
In Table~\ref{tab:abstudy1}, `$n$ SA Modules' means the transformer of our basic network has $n$ self-attention modules.
Comparing `2 SA Modules', `4 SA Modules', and SiamTrans that uses 6 modules, we find that more self-attention modules can get better results.

\textbf{Transformer or CNN.}
We replace the transformer with the UNet~\cite{ronneberger2015u} which has the same model size with the transformer in the basic network (denoted as `UNet' in Table~\ref{tab:abstudy1}). Two feature maps outputed from the encoders are concatenated and served as the input of the UNet. The result shows that using the transformer is better than using the UNet in the basic network. 

\subsection{Practical Applications of our Method}
From the above experiments, our multi-frame method performs better than previous single- or multi-frame methods.
It requires to use multiple consecutive frames.
In practice, it is easy to obtain multiple frames from modern cameras or mobile phones.
These equipments have the burst mode and we can get multiple images from one scene in a short period.
These multiple frames can also be extracted from videos, like the \textit{NTURainReal} dataset we obtain in Sec. \ref{sec:dataset}.

\section{Conclusions}
In this paper, we have proposed a zero-shot multi-frame image restoration method for removing unwanted obstruction elements that vary in successive frames.
Our method contains three stages.
After self-supervisedly pre-trained on the denoising task, our SiamTrans model is tested on three tasks unseen during the pre-training.
The results of SiamTrans are further improved by the hard patch refinement.
Compared with a number of supervised or unsupervised, single-frame or multi-frame state-of-the-arts, our method achieves the best performance on all the tasks and on all the datasets.
The future work includes applying SiamTrans to other tasks to explore more of its capacity.

		\bibliography{aaai22}

\newpage
\appendix
\onecolumn
\section{The structures of the CNN encoder and decoder}
\begin{figure*}[h]
		\centering
		\includegraphics[width=0.78\textwidth]{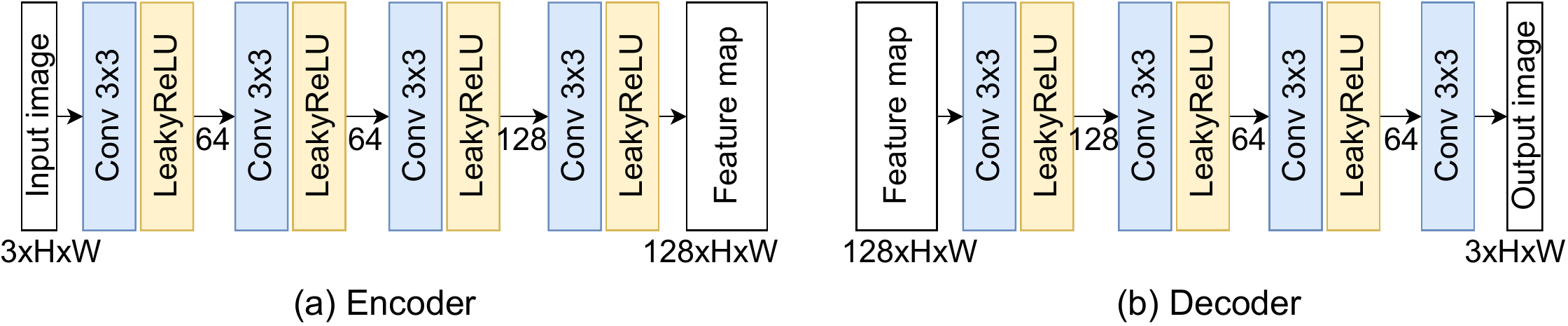}
		\caption{The structures of the encoder (a) and decoder (b) in the basic network. Both of them contain four $3 \times 3$ convolutional layers.}
		\label{network_encoder}
\end{figure*}

\section{Synthesizing rain sequences and snow sequences for supervised methods}
\begin{figure*}[h]
		\centering
		\includegraphics[width=0.88\textwidth]{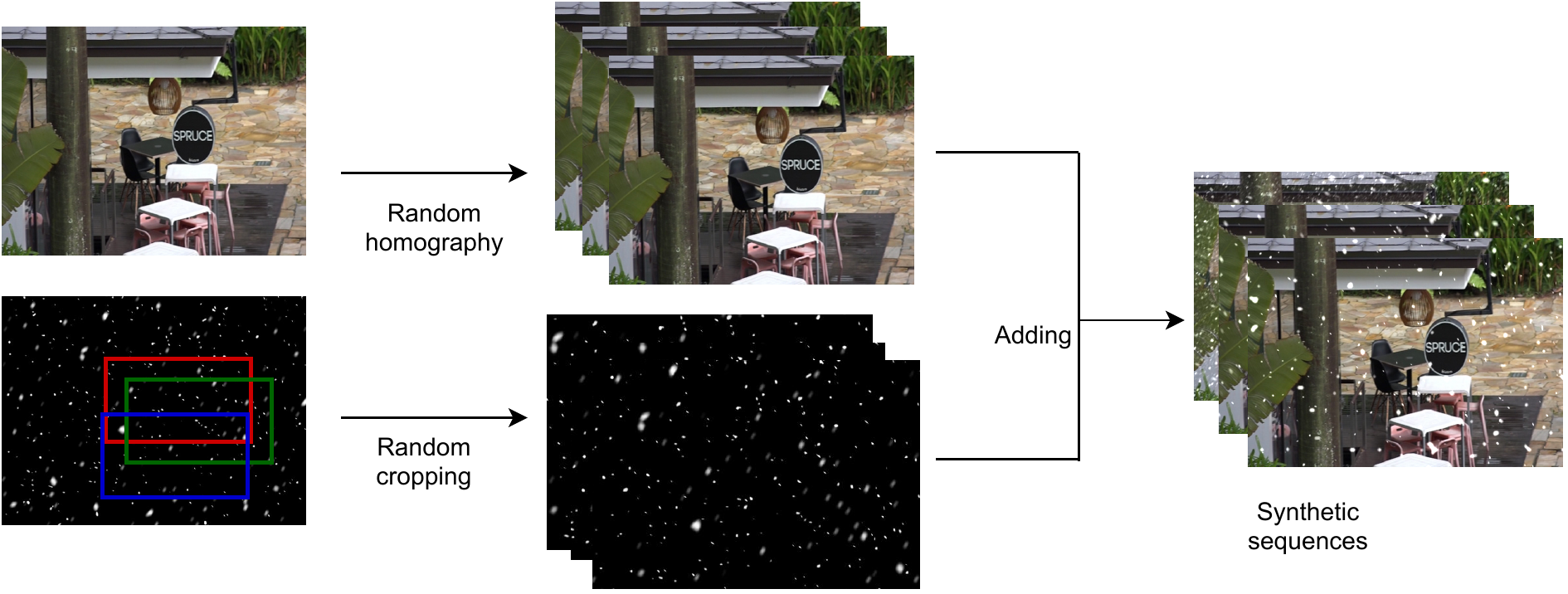}
		\caption{The generation process of synthetic snow sequences for the supervised methods.}
		\label{fig:pro_syn}
\end{figure*}

Since there is no large scale multi-frame deraining dataset or desnowing dataset, we synthesize training data for supervised methods for comparison.
We illustrate the process of the synthetic snow sequences in Fig.~\ref{fig:pro_syn}. Random homography and cropping are firstly applied to the clean and snow images respectively, and then the transformed images and the snow croppings are added together to obtain the sequence frames. The same process is applied to obtain rain sequences.

\section{Examples of the multi-frame datasets}
\begin{figure}[h!]
		\centering
		\includegraphics[width=0.88\textwidth]{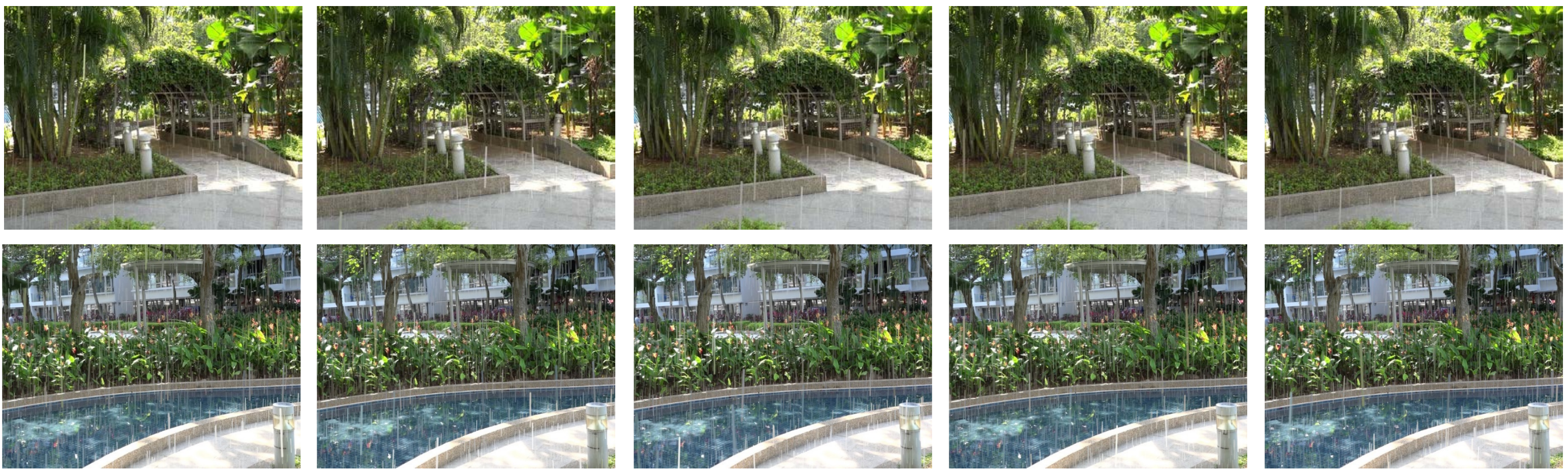}
		\caption{Some examples of \textit{NTURain}.}
		\label{fig:rain2}
\end{figure}

\begin{figure}[h!]
		\centering
		\includegraphics[width=0.88\textwidth]{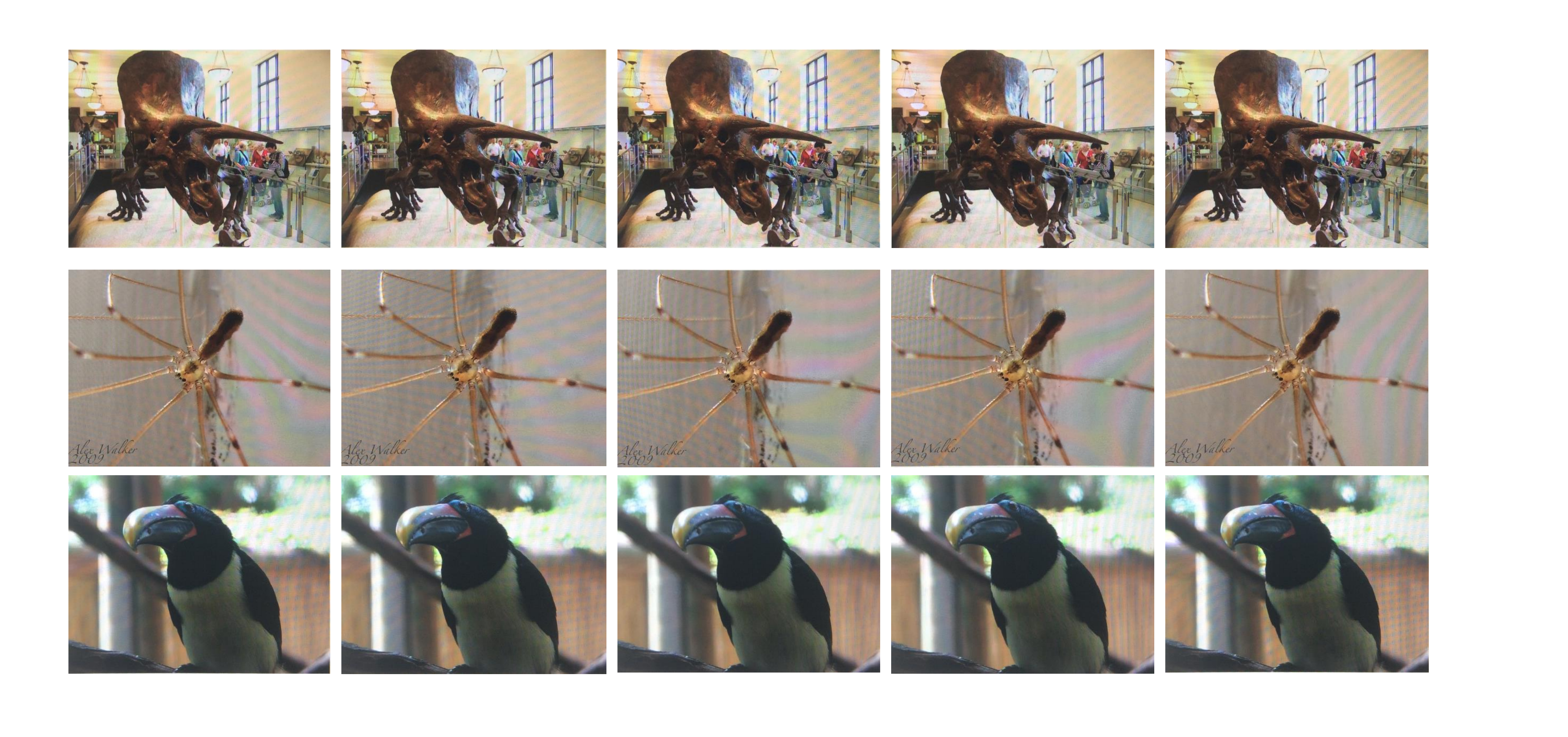}
		\caption{Some examples of \textit{MFMoire}.}
		\label{fig:moire2}
\end{figure}

\begin{figure}[h!]
		\centering
		\includegraphics[width=0.88\textwidth]{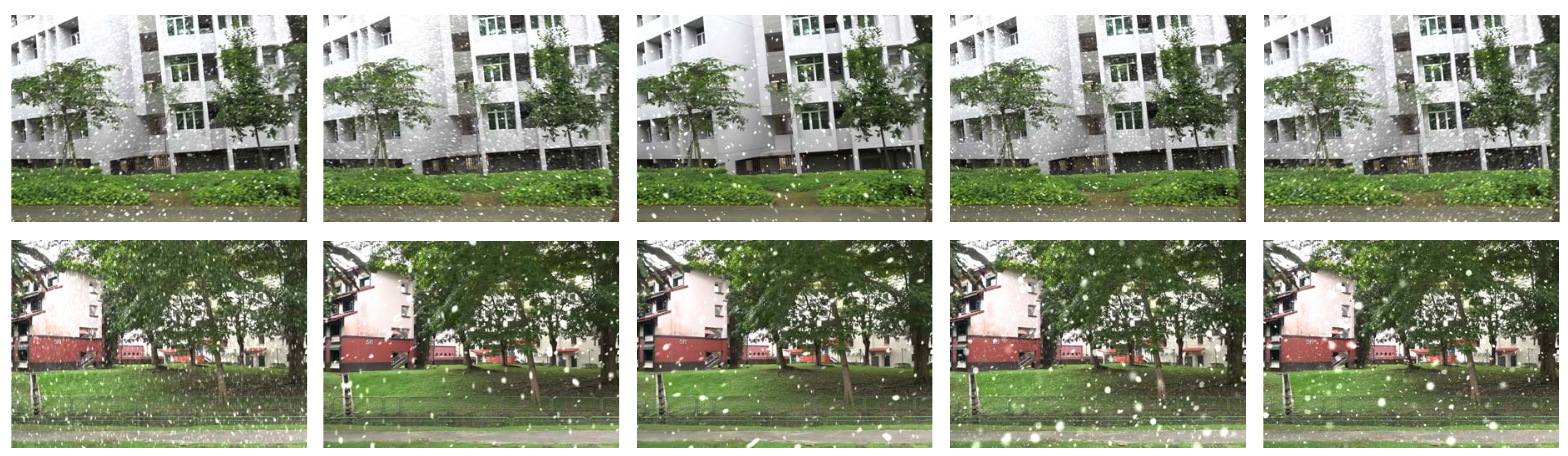}
		\caption{Some examples of \textit{MFSnow}.}
		\label{fig:snow}
\end{figure}

Fig.~\ref{fig:rain2}, Fig.~\ref{fig:moire2}, and Fig.~\ref{fig:snow} show some examples of \textit{NTURain}, \textit{MFMoire}, and \textit{MFSnow}, respectively. Note that though the time intervals between consecutive shots are very short, the rain/snow/moire patterns vary a lot.

\clearpage
\section{Comparison with single-frame methods}
\begin{figure}[h]
		\centering
		\includegraphics[width=0.85\textwidth]{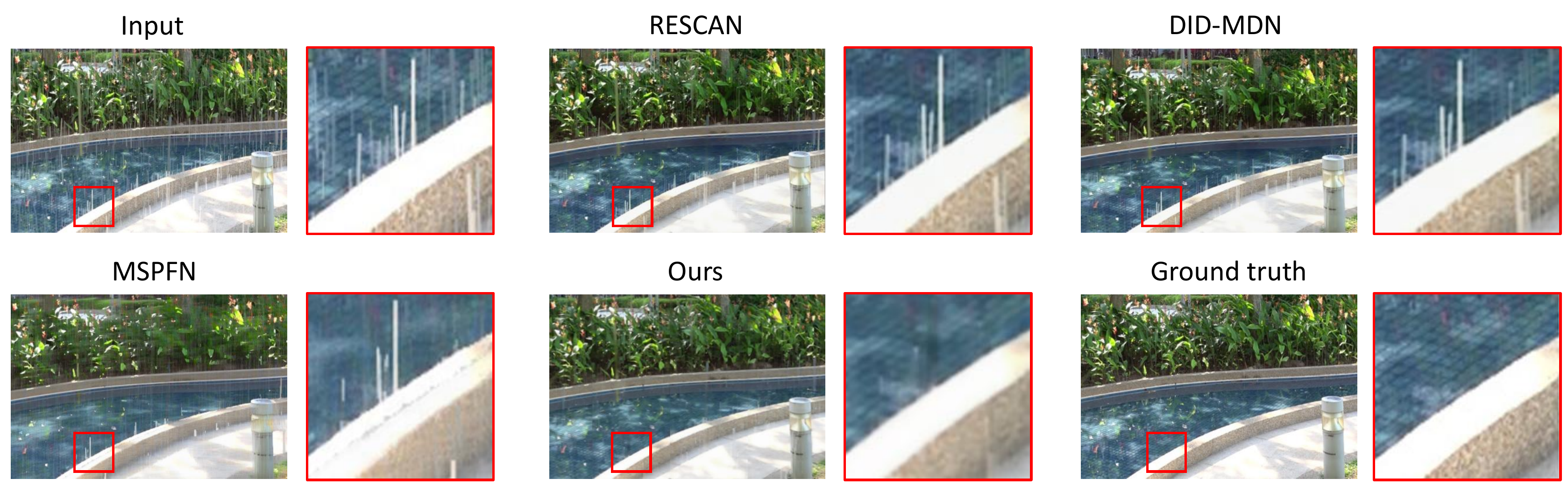}
		\includegraphics[width=0.85\textwidth]{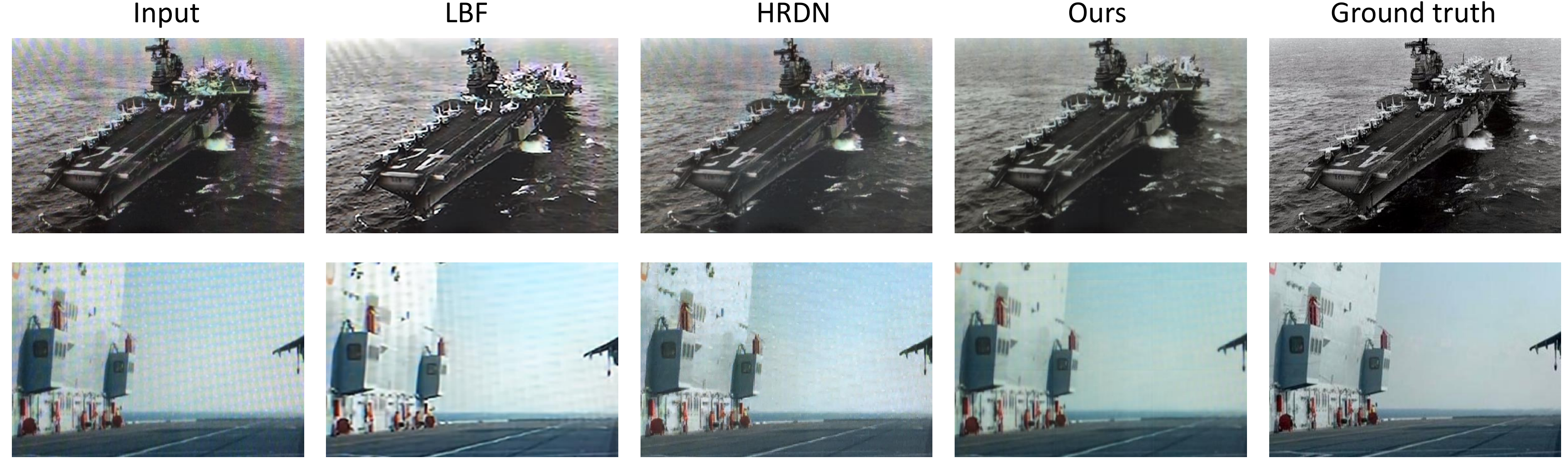}
		\caption{\textbf{Top}: visual deraining comparison among ours and three other methods (RESCAN~\cite{li2018recurrent}, DID-MDN~\cite{derain_zhang_2018}, and MSFPN~\cite{Kui_2020_CVPR}). \textbf{Bottom}: Visual demoireing comparison among ours and two single-frame demoireing methods (LBF~\cite{zheng2020image} and HRDN~\cite{yang2020high}).}
		\label{fig:derain_sig}
\end{figure}

In Table~\ref{tab:nturain2} and Table~\ref{tab:mfmoire2}, we compare our method with several state-of-the-art single-frame methods. Ours outperforms these methods by large margins. Fig.~\ref{fig:derain_sig} shows our method (based on multiple frames) can remove the rain or moire patterns more thoroughly.
 \begin{table}[h]
  \begin{center}
  \caption{Quantitative deraining comparison. The best results are in \textbf{bold}. }
  \label{tab:nturain2}
  \resizebox{12.5cm}{!}{
  \begin{tabular}{cccccc }
    \toprule
      Method & \ &RESCAN~\cite{li2018recurrent}  &MSPFN~\cite{Kui_2020_CVPR}&DID-MDN~\cite{derain_zhang_2018} & SiamTrans (Ours)\\
    \midrule 
    \multirow{2}*{NTURainSyn}&PSNR$\uparrow$&25.80 &25.16&25.48&$\mathbf{27.02}$\\
    &SSIM$\uparrow$& 0.8716&0.8497&0.8838&$\mathbf{0.8991}$\\
    \midrule 
    NTURainReal&NIQE$\downarrow$&3.437&3.462 &3.416&$\mathbf{3.302}$\\
    
    \bottomrule
  \end{tabular}}
  \end{center}
\end{table}

\vspace{-0.6cm}

\begin{table}[h]
  \begin{center}
  \caption{Quantitative demoireing and desnowing comparison. The best results are in \textbf{bold}. }
  \label{tab:mfmoire2}
  \resizebox{7cm}{!}{
  \begin{tabular}{ccccc }
    \toprule
      &Method & PSNR$\uparrow$ & SSIM$\uparrow$ &LPIPS$\downarrow$\\
    \midrule 
    \multirow{4}*{MFMoire}&LBF~\cite{zheng2020image}&19.78&0.6153& 0.3892\\
    &MopNet~\cite{he2019mop}&19.91&0.6165& 0.3320\\
    &HRDN~\cite{yang2020high}&20.39&0.6215 &0.3269\\
    &SiamTrans (Ours)&$\mathbf{22.26}$&$\mathbf{0.6642}$ &$\mathbf{0.3197}$ \\
    \midrule
    \multirow{2}*{MFSnow}&JATASR~\cite{chen2020jstasr}&21.06&0.7309&0.2382 \\
    &SiamTrans (Ours)&$\mathbf{26.05}$&$\mathbf{0.8605}$ & $\mathbf{0.1323}$\\
    \bottomrule
  \end{tabular}}
  \end{center}
\end{table}
\vspace{-0.6cm}

\clearpage
\section{Visual desnowing comparison among our method and other algorithms}
\begin{figure}[h]
		\centering
		\includegraphics[width=0.94\textwidth]{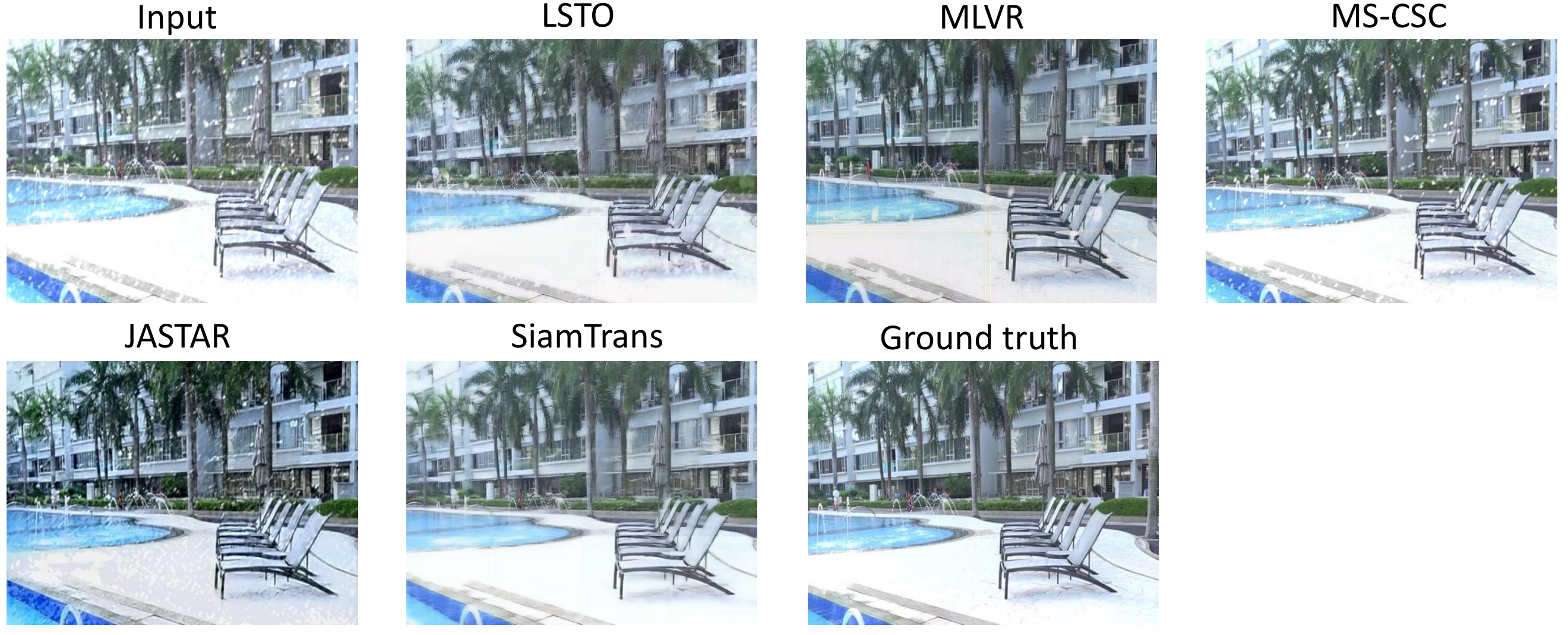}
		\caption{Visual desnowing comparison among our method and four other algorithms (LSTO~\cite{liu2020learning}, MLVR~\cite{alayrac2019visual}, MS-CSC~\cite{li2019video}, and JASTAR~\cite{chen2020jstasr}).}
		\label{fig:desnow}
\end{figure}

\section{The procedure of the refinement in the stage of hard patch refinement (HPR)}

 \begin{algorithm}[H]  
  \caption{The procedure of the refinement.}  
  \label{alg:Framework2}  
  \begin{algorithmic}[1]  
    \REQUIRE   
    $n \times j$ patches ${p^{1}_{j},p^{2}_{j},...,p^{n}_{j}}$ ~$(j=1,2,...,N)$;  
    \ENSURE  
    Update these patches ${p^{1}_{j},p^{2}_{j},...,p^{n}_{j}}$~ $(j=1,2,...,N)$;  
           \FOR{ $iter$ = 1 to $20$}
           \FOR{ $k$ = 1 to $n$}
    \STATE $p^{k}_{1} \leftarrow \ \alpha \times p^{k}_{1} + \frac{ 1 - \alpha}{N-1} \!\left(\sum_{m=2}^{N} \!W_{1}(p^{k}_{m}\!\uparrow\!)\right)\!\downarrow$;
           \ENDFOR
           \FOR{ $j$ = 2 to $N$}
           \FOR{ $k$ = 1 to $n$}
    \STATE $p^{k}_{j} \leftarrow \  \alpha \times p^{k}_{j} + \frac{ 1 - \alpha}{N-2} \!\left(\sum_{m=2, m\neq j}^{N} \!W_{j}(p^{k}_{m}\!\uparrow\!)\right)\!\downarrow$;
           \ENDFOR
           \ENDFOR
           \ENDFOR
    \RETURN 
  \end{algorithmic}  
\end{algorithm}   
\end{document}